\pdfoutput=1
\documentclass{bmvc2k}
\usepackage{booktabs,siunitx,caption}
\usepackage[colorinlistoftodos,prependcaption,textsize=tiny]{todonotes}
\usepackage{multirow}
\usepackage{makecell}
\usepackage{amsfonts}
\usepackage{floatrow}
\usepackage{pifont}
\usepackage{paralist}
\usepackage{bm}
\usepackage{enumitem}
\usepackage{soul}
\usepackage{wrapfig}

\newcommand{\method}{\texttt{TADeT}\xspace}
\newcommand{\ba}{{Balanced Accuracy}\xspace}
\newcommand{\sa}{{Standard Accuracy}\xspace}
\newcommand{\eo}{{Equalized Odds}\xspace}
\newcommand{\bad}{{Balanced Accuracy Difference}\xspace}

\newcommand{\img}{{\bm{x}}\xspace}
\newcommand{\patch}{{x}\xspace}
\newcommand{\lbl}{{y}\xspace}
\newcommand{\attr}{{a}\xspace}
\newcommand{\numlayer}{{L}\xspace}
\newcommand{\encmodel}{{f}\xspace}
\newcommand{\clsmodel}{{h}\xspace}
\newcommand{\discrmodel}{{g}\xspace}
\newcommand{\query}{{Q}\xspace}
\newcommand{\key}{{K}\xspace}
\newcommand{\val}{{V}\xspace}
\newcommand{\channel}{{D}\xspace}
\newcommand{\attnI}{{m}\xspace}
\newcommand{\channelI}{{d}\xspace}
\newcommand{\numHead}{{M}\xspace}
\newcommand{\batch}{{B}\xspace}

\title{Mitigating Bias in Visual Transformers via Targeted Alignment}

\addauthor{Sruthi Sudhakar}{sruthis@gatech.edu}{1}
\addauthor{Viraj Prabhu}{virajp@gatech.edu}{1}
\addauthor{Arvindkumar Krishnakumar}{akrishna@gatech.edu}{1}
\addauthor{Judy Hoffman}{judy@gatech.edu}{1}

\addinstitution{
 Georgia Institute of Technology\\
 Atlanta, GA
}

\runninghead{Sudhakar et al.}{Mitigating Bias in Visual Transformers}

\begin{document}

\maketitle
\vspace{-14pt}
\begin{abstract}
\vspace{-3pt}

As transformer architectures become increasingly prevalent in computer vision, it is critical to understand their fairness implications. We perform the first study of the fairness of transformers applied to computer vision and benchmark several bias mitigation approaches from prior work. We visualize the feature space of the transformer self-attention modules and discover that a significant portion of the bias is encoded in the query matrix. With this knowledge, we propose \method, a targeted alignment strategy for debiasing transformers that aims to discover and remove bias primarily from query matrix features. We measure performance using \ba and \sa, and fairness using \eo and \bad. \method consistently leads to improved fairness over prior work on multiple attribute prediction tasks on the CelebA dataset, without compromising performance.

 %\todo[inline]{code link}.
\end{abstract}
\vspace{-0.7cm}
\section{Introduction}
\label{sec:intro}
\vspace{-0.2cm}

Over the past decade, deep learning-based computer vision has been advancing at a rapid pace and is being deployed in high-stakes applications ranging from candidate job hiring to facial recognition systems. Several recent studies have shown that deep models take advantage of spurious correlations between attributes in the data when learning to make predictions. Such unintended correlations can cause the model to act in a biased way, such as having lower accuracy on certain sub-populations of the data \cite{gendershades,singh2020dont,willson2019obj,kay2015unequal, du2020fairness}. These biases can be harmful and cause discrimination towards these populations once deployed. 

While the fairness of computer vision tasks learnt using convolutional neural network (CNN) models is well explored, the fairness implications of \emph{Visual Transformers} are yet to be studied. Transformers were first introduced in NLP \cite{vaswani2017attention}, and later found to be effective on computer vision tasks \cite{dosovitskiy2020image} due to their ability to learn long-range relationships, unlike CNN’s which have have a limited field of view at any given layer. There has been a rise in the use of transformers in vision tasks~\cite{khan2021transformers}. In this paper, we perform the first study of visual transformers with respect to fairness. Transformers have a naturally separated and spatially cohesive feature space \cite{vaswani2017attention}, and we investigate how this unique architecture may be leveraged to ensure fairness. 

There are several ways to measure the fairness of a model~\cite{dwork2011fairness, hardt2016equality, zhao2017men, dieterich2016compas}.
%including: Demographic Parity \cite{dwork2011fairness}, Equality of Opportunity \cite{hardt2016equality}, \eo \cite{hardt2016equality}, Bias Amplification \cite{zhao2017men}, and Accuracy Equity \cite{dieterich2016compas}. While Demographic Parity ensures equal positive rates across a protected group, it does not maintain the accuracy of the system, rendering it impractical for most practitioners. Furthermore, both Demographic Parity and Equality of Opportunity only focus on the positive outcome in binary classification tasks, however, many settings do not have a 'positive' outcome that is favored, rather, fairness across both outcomes is preferred. Bias Amplification does not address underlying dataset bias. Therefore, 
In this work, we focus on an image classification task for which all predictive outputs are equally desirable, and hence measure performance using \emph{balanced} (multiclass) accuracy~\cite{park2020readme}. Further, we focus on two fairness metrics: \eo~\cite{dwork2011fairness}, as it focuses on equal true positive and false positive rates across the protected attribute, and a new auxiliary measure we introduce, \textbf{\bad} ($\Delta$ BA) which measures the difference in \ba across the two groups. This metric, which is similar to accuracy equity~\cite{dieterich2016compas}, gives practitioners an overall idea of the \emph{difference} in model performance across a protected attribute. 
%We measure performance of our model using \ba \cite{park2020readme} and \sa.

Several debiasing algorithms have been proposed and studied for CNNs, including re-weighting instances~\cite{wang2019balanced}, MMD alignment~\cite{long2015learning}, domain independent training~\cite{dwork2011fairness,wang2020fairness}, and adversarial training~\cite{zhang2018mitigating}. However, many of these techniques suffer from a trade-off where increased bias mitigation results in reduced predictive performance~\cite{wang2020fairness}. In addition, these works either require a large number of the protected class data or are designed for learning with a CNN model, making it unclear how well these methods will translate to newer backbone architectures like visual transformers. 

%Firstly, re-weighting or re-sampling data was proven to not work in\cite{dwork2011fairness} all situations, even when the system was fed with more data \cite{wang2019balanced}. Next, MMD alignment is to simplistic as it simply aims to match the distributions \cite{madras2018learning}. Domain training was proposed in \cite{dwork2011fairness} and followed up in \cite{wang2020fairness}, however this method lacks robustness when certain classes are severely underrepresented in the data and leads to drops in accuracy. Finally, adversarial debiasing has been shown to work well, but often with drops in accuracy \cite{wang2020fairness}.
\begin{figure}[t]
\centering
\includegraphics[width=0.65\textwidth]{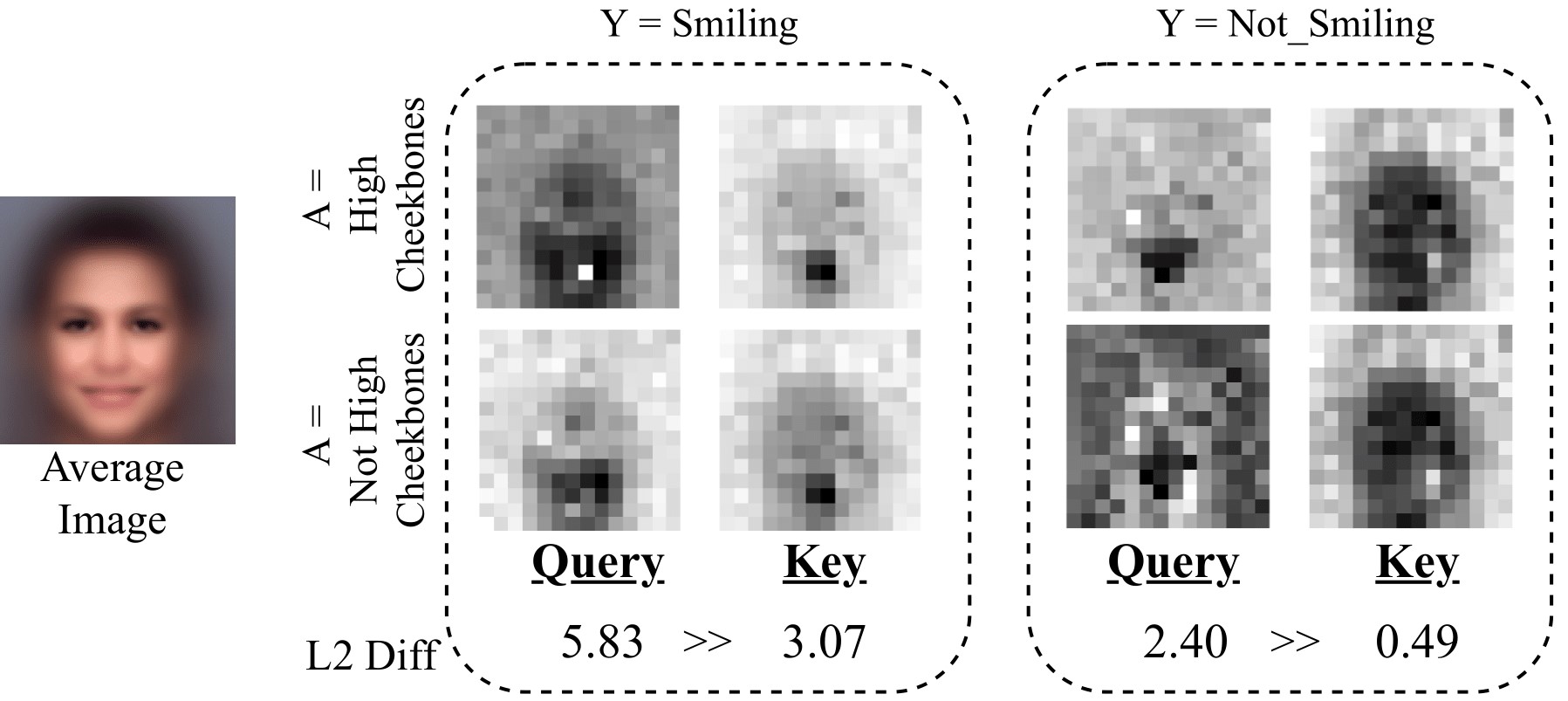}
\caption{We discover that bias in a visual transformer model across a particular attribute label (e.g. A = high cheekbones / not high cheekbones) can be attributed to a variance in the query matrix activations for a fixed task label (Y = smiling / not smiling), even with similar key activations. This observation motivates \method (example shown on CelebA~\cite{liu2015deep}).}
%have high variance across an attribute label (A = High Cheekbones or not High Cheekbones) on the CelebA dataset. We find that the difference across attribute labels is larger for query than key matrices regardless of the output task label. This indicates possible bias localized in Query Matrices.}
\label{fig:qmbefore}
\vspace*{-5pt}
\end{figure}

%While several studies exist on CNNs, 
We perform the first fairness assessment of visual transformer models and benchmark existing bias mitigation algorithms. 
We then introduce \textbf{\method}, Targeted Alignment for Debiasing Transformers, a novel algorithm that applies class-specific Maximum Mean Discrepancy (MMD)~\cite{gretton2008kernel} alignment and class-specific adversarial~\cite{zhang2018mitigating} debiasing, to a \emph{targeted} part of the visual transformer feature space. We discover that when bias exists for a particular attribute, the core piece of the model which differs between two subgroups (one with and one without said attribute) can be observed through the query activations. For example, in Figure~\ref{fig:qmbefore} we consider a Transformer trained to predict the presence of the ``Smiling'' attribute, and visualize its \emph{average} query and key matrix activations over the dataset. We find that for the subset of images where smiling is present, a large discrepancy exists in average query activations across the protected attribute of ``High cheekbones'' (box 1, column 1), whereas the key activations are much more similar (box 1, column 2). This result is also observed in the case of \emph{not} smiling images (box 2). 
% While it is not necessary for the query and key activations to align with each other, we would expect the average activataions for (smiling, high cheekbone) and (smiling, not high cheekbone), to be more similar for both queries for keys
Thus, by imposing alignment~\cite{ganin2015unsupervised} directly on the query activations we may be able to effectively reduce model bias while limiting the impact on predictive performance.
%approaches and uses specific features of the Transformer model architecture to improve fairness metrics while maintaining model accuracy. 

We evaluate our method on the task of single-label attribute prediction on the CelebA dataset \cite{liu2015deep}, while requiring invariance to a defined protected attribute. We measure the performance of our method using \ba and \sa, and the fairness of our method using \eo \cite{hardt2016equality} and \bad.
We make the following contributions:
\begin{compactenum}
  \itemsep0em 
  \item We perform the first assessment of the fairness of representations learned by visual transformers, and benchmark several existing debiasing methods from the literature.
  \item We visualize the transformer feature space and discover that a significant portion of learned bias is encoded in the activations of the transformer's query matrices.
  \item Motivated by this, we propose \method, a debiasing strategy for visual transformers using \emph{targeted}, \emph{class-specific}, MMD alignment and adversarial debiasing to improve upon existing fairness algorithms without compromising accuracy.
\end{compactenum}

\vspace{-0.5cm}
\section{Related Work}
\vspace{-0.3cm}
\label{sec:relwork}

\noindent\textbf{Transformers in computer vision.}
Transformers were first introduced in NLP by Vaswani~\emph{et al.}~\cite{vaswani2017attention}, which used self-attention to develop better context and allow models to attend to relevant words across the entire corpus of text. Recently, ViT (Visual Transformer)~\cite{dosovitskiy2020image} has emerged as a new architecture for vision capable of achieving state-of-the-art performance on standard recognition tasks. Given the strong performance of this architectural paradigm, many future works are bound to leverage the visual transformer model, necessitating a study into the fairness implications and mitigation approaches designed for transformers. 
%Given the recency of this work no studies have yet been conducted to measure the fairness implication of these architectures. the promise of such approaches 
%There has been no work that we know of which analyzes where the bias is encoded in Transformers. Therefore, we perform an analysis of the feature space, specifically the Query matrices, of the Transformer model to understand where the bias is being encoded and how to design an algorithm to mitigate this bias. 

\noindent\textbf{Measuring Biases.}
Several metrics exist for measuring bias in machine learning models.
Demographic parity~\cite{dwork2011fairness} measures whether the rates of positive outcomes across all protected groups are equal, making it susceptible to the self-fulfilling prophecy and subset targeting.
%demonstrates a few examples where this metric fails, with the most concerning one being the self-fulfilling prophecy and subset targeting. 
%\jh{Can you comment on how this is specific to the positive outcome as opposed to balanced approaches.}
%Self-fulfilling prophecy is the fear that members belonging to a protected group are given the positive outcome more often solely to equalize the positive rate, causing disregard for the ground truth because it does not require the model to make accurate predictions. This can result in members from a protected attribute being incorrectly given the positive outcome a greater number of times which can increase the bias.
Bias Amplification~\cite{zhao2017men} measures how a model \emph{amplifies} biases, though it does not provide feedback about the inherent biases common in visual recognition datasets \cite{tommasi2015deeper, bhargava2019exposing, wang2019balanced, denton2020image}.
\eo~\cite{hardt2016equality}, one of the metrics we report, offers a stricter definition of Equality of Opportunity, by adding an extra constraint on the Demographic Parity measure where each group must receive the positive and negative outcomes at equal rates if the datapoints in this group qualify for that outcome. 
%We choose to use \eo as one of our metrics as it focuses on both model and dataset bias (unlike Bias Amplification) and resolves some issues with self-fulling prophecy in Demographic parity. 
%Finally, we introduce Balanced Accuracy Difference, a metric for capturing the overall performance gap 
%model exhibits on a subpopulation compared to the rest of the data. 
Finally accuracy equity \cite{dieterich2016compas} compares accuracy across subpopulations. However, when a dataset is class imbalanced, this measurement will have an implicit bias against the minority class. Therefore, we introduce Balanced Accuracy Difference as a way to measure the difference in a model's performance across a protected attribute while accounting for class imbalance.

\noindent\textbf{Mitigation Methods.}
Some of the most straightforward mitigation methods focus on re-sampling or re-weighting techniques~\cite{elkan2001foundations,bickel2009discriminative}, but these methods may have limited information gain through resampling and in some cases suffer from overfitting~\cite{Weiss2007CostSensitiveLV}.
A different style of approach seeks to directly impose alignment of feature spaces by optimizing an MMD loss between features from different subgroups~\cite{long2015learning}, though mitigation benefits are variable~\cite{madras2018learning}. 
A domain discriminative~\cite{dwork2011fairness} approach trains a classifier for each protected attribute and task combination and hence attempts to induce 'fairness through awareness', though recent works~\cite{wang2020fairness} demonstrate limitations including a lack of leveraging between-class information and an increased number of parameters.
A domain independent approach~\cite{wang2019balanced} trains an ensemble of classifiers with a shared feature representation to ensure that rare attribute values are adequately represented. While this approach can lead to large bias reductions, it is limited in that it assumes adequate labeled data for each setting of the protected attribute in order to train a representative model. 

%\noindent \textbf{Domain Independent}~\cite{wang2020fairness}: This method improves upon Domain Discriminative training by using one classifier per protected attribute, and combining these predictions during post processing to ensure the protected domains are weighted fairly. This reduced computation costs greatly while still achieving similar results. However, a major cause for concern is that if there are not many instances of a protected attribute, training 1 classifier explicitly per domain could lead to further unintended biases. Furthermore, there exists an extra cost of training a separate classifier per protected attribute. Finally, we observe that despite prior claims, this method leads to decreased equalized accuracy and normal accuracy compared to adversarial learning. Due to these pitfalls, we follow the 'fairness through unawareness approach'.

Many recent approaches have leveraged the success of adversarial learning techniques for domain alignment~\cite{ganin2015unsupervised}. Adversarial Debiasing~\cite{zhang2018mitigating} alternates between training a classifier to distinguish the protected attribute and updating the representation to be invariant to the protected attribute. 
%which sets up a min-max game between two players, where the main classifier tries to achieve its task of correctly predicting the labels, and meanwhile an adversary tries to extract information about a protected attribute. The goal is for the classifier to learn a representation that contains less information about the protected attribute. This method was proven to be very successful, however limitations include small drops in accuracy, lack of stability, and difficulty in tuning hyper-parameters.
LAFTR~\cite{madras2018learning} is an adversarial method with the ability to modify the training objective based on the desired fairness measure (equalized odds, equality of opportunity, and demographic parity). However, in our experiments using this method with the equalized odds objective, we find that it is unable to consistently achieve fairness improvements while maintaining high performance.

We benchmark MMD, Domain Independent, Adversarial Debiasing, and LAFTR methods on transformers. We then propose \method, a targeted alignment approach to debias transformers (leveraging the query matrix features) that achieves improved fairness with minimal reduction in predictive performance. 

\vspace{-0.5cm}
\section{Approach}
\vspace{-0.3cm}
Recently, visual transformers~\cite{dosovitskiy2020image} have emerged as a popular architecture for learning visual recognition models. While the fairness of CNN models has been documented~\cite{gendershades} and mitigation strategies have been proposed~\cite{zhang2018mitigating,madras2018learning}, we study the bias which is present within a visual transformer-based model and propose \method, a new bias mitigation algorithm designed for debiasing visual transformers. \method focuses alignment on query activations within the transformer feature space to remove bias against a protected attribute.

\vspace{-.4cm}
\subsection{Notation and Background}
\vspace{-0.2cm}
Let $\img$ denote an input image and $\lbl$ the corresponding task label. Our goal is to learn a classifier $\clsmodel$, on top of deep encoder, $\encmodel$, such that $\clsmodel(\encmodel(\img)) \xrightarrow{} \lbl$ 
%parameterized by $\theta$, -- REMOVING THIS SINCE YOU DON'T USE IT
 achieves high heldout task accuracy while being fair with respect to a protected attribute, $\attr$. Fairness has many definitions; in this work we focus on designing a model which is invariant with respect to the protected attribute (see Sec~\ref{sec:relwork}). 

The model $\encmodel$ employs the visual transformer architecture introduced in Dosovitskiy~\emph{et al.}~\cite{dosovitskiy2020image}. The input image $\img \in \mathbb{R}^{H \times W \times C}$ is reshaped into $N$ patches $\{\patch_1, .., \patch_N \}$, where each patch $x_i$ is of size $\frac{HW}{N} \times \frac{HW}{N} \times C$. Each patch is then flattened and mapped to $\channel$ dimensions with a trainable linear projection. A final learnable class token $\patch_{\text{class}}$ of the same dimension, $\channel$, is appended to the sequence of patches. This class token is randomly initialized and updated throughout training.
These $N+1$ patch embeddings are appended with a positional embedding to retain positional information, and then fed through $\numlayer$ layers of the transformer encoder. Each layer of the encoder is identical and includes a multi-headed self-attention module followed by alternating MLP blocks and layernorms. 

The multi-headed self-attention module includes $\numHead$ attention heads. For each attention head, $\attnI$, the concatenated $N+1$ patch embeddings are split into 3 branches and each fed to a linear layer. The 3 branches result in learning Query ($\query$), Key ($\key$), and Value ($\val$) activations respectively, each of dimension $\numHead \times (N+1) \times \channel$, which attempt to encode specific pieces of information. The value matrix is tasked with learning an encoding of the actual attributes of the data (\emph{eg.} for a face image, information such as hair color, eye color, etc.). The key matrix encodes the location of \emph{where} specific information is stored in the value matrix (\emph{eg.} where the eye color is located). Finally, the query matrix is trained to ``ask'' the key matrix for the information it requires for the task. Activations produced from these 3 matrices are then combined to produce the self-attention defined in Vaswani~\emph{et al.}\cite{vaswani2017attention} as:
\begin{equation} \label{eu_eqn}
\text{Attention}(\query, \key, \val) = \text{softmax}\left(\frac{\query \key^T}{\sqrt{\channel}}\right)\val
\end{equation}
The activations from the self-attention module are fed into alternating MLP and layernorms in each layer of the encoder. Finally, the features for the class token, $\patch_{\text{class}}$, are extracted from the output of the final layer transformer encoder and used as the feature representation of input $\img$ to predict output $\lbl$. 
\label{sec:approach}

\begin{figure}[t]
\centering
\includegraphics[width=0.6\textwidth]{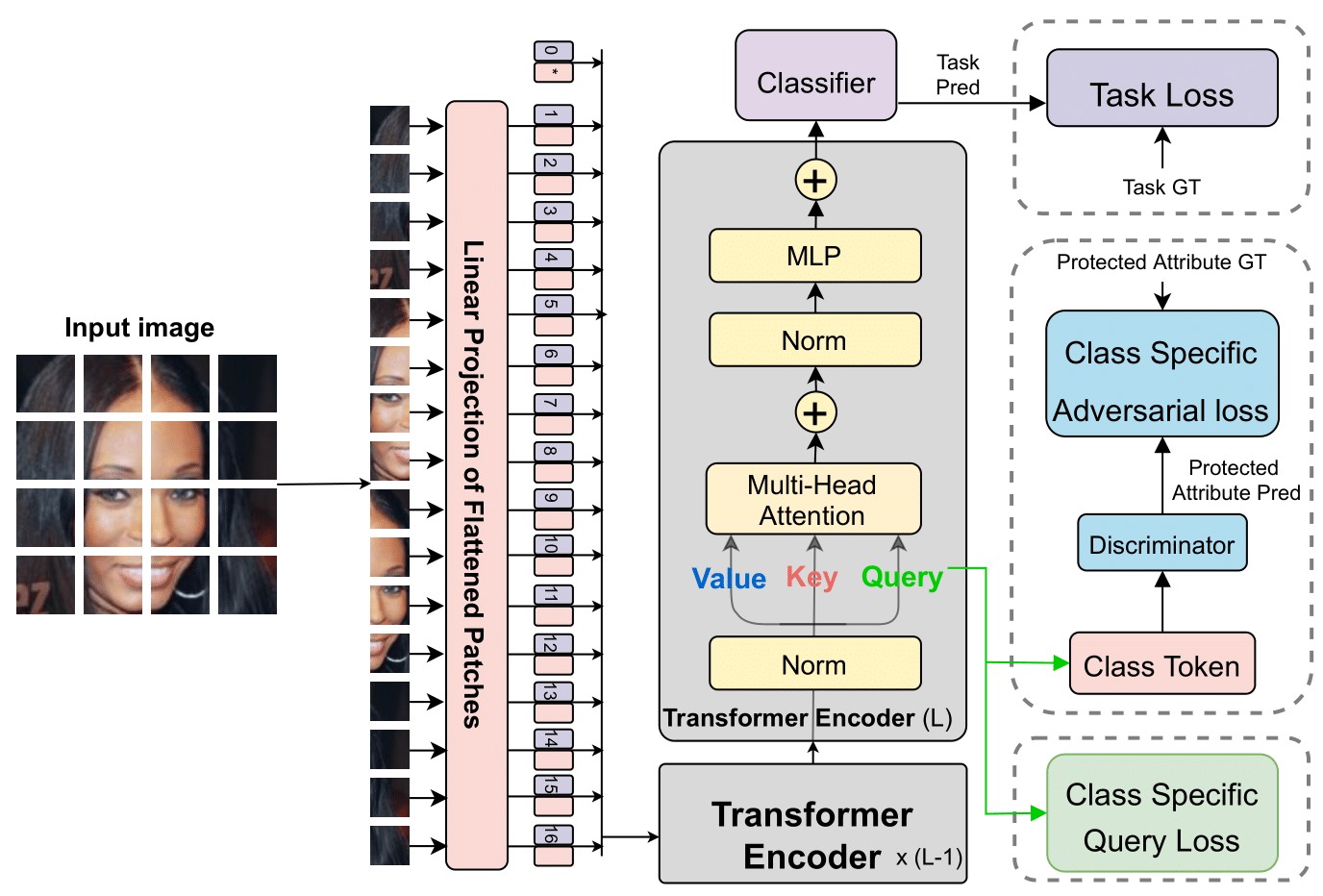}
\vspace{-10pt}
\caption{Overview of \method. First, a preprocessed input image and an additional class token are passed through $\numlayer$ transformer encoder layers, and features extracted from the class token are used for task classification. Second, query activations of the last layer of the encoder are used for Targeted Debiasing in two ways: i) query activations corresponding to the class token are used for class-specific adversarial alignment (Sec.~\ref{subsec:advloss}), and ii) for each class, the $L_2$ distance between average query activations for each setting of the protected attribute is minimized (Sec.~\ref{subsec:queryloss}).}
\label{fig:arch}
\end{figure}

\vspace{-.3cm}
\subsection{\method: Targeted Alignment for Debiasing Transformers}
\label{subsec:tadot}
\vspace{-0.2cm}
Unlike CNNs, transformers possess a naturally interpretable feature space due to separation into queries and keys. On studying the average query and key activations of the last layer of a visual transformer encoder before making a prediction, we find a measurable difference across average query activations across the protected attribute values, but only minimal differences between average key activations across the protected attribute values (see Figure~\ref{fig:qmbefore}). These differences generalize across the D dimensions and M attention heads of the query matrix. This observation of a large difference in query activations motivates our focus on alignment of query activations.

We introduce \method, Targeted Alignment for Debiasing Transformers, a bias mitigation algorithm that focuses on aligning \emph{query activations} (see Figure~\ref{fig:arch}). \method considers a visual transformer model trained using a standard task loss and performs debiasing through the introduction of two class-specific alignment losses, 1) an adversarial loss which aligns the class token of the query activations, $L_\text{adv}$, (Sec~\ref{subsec:advloss}) and, 2) an $L_2$ loss which penalizes large deviations between the average query activations for values of the protected attribute, $L_q$, (Sec \ref{subsec:queryloss}). Our overall model considers the joint optimization between these two objectives together with a task supervised cross-entropy loss, $L_\text{CE}$.
\vspace{-0.6cm}
\subsubsection{Class-specific Adversarial Alignment}
\label{subsec:advloss}
\vspace{-0.2cm}
We begin alignment across populations with different attributes by leveraging adversarial mitigation~\cite{zhang2018mitigating}.
In adversarial training for feature alignment, an adversary $\discrmodel$ is applied to a feature space to determine if two groups are distinguishable. Optimization alternates between training the discriminator to distinguish the groups, $\attr$ in our case, and updating the feature encoder, $\encmodel$, to confuse the discriminator. Whereas prior work aligns the penultimate features for an instance, $\img$, we seek to align the class token of the query matrix, $\query_\img[:,N+1,:]$. This results in a standard min/max optimization over this targeted feature space:
%where the discriminator is trained to predict the attribute label with a cross-entropy loss and the encoder is optimized to maximize the cross-entropy loss:
\begin{equation} \label{eq:minmax}
\mathcal{L}_\text{adv}(\img; \encmodel, \discrmodel) = \max_{\encmodel}\min_{\discrmodel}\mathbb{E}_{\img,\lbl,\attr}[\mathcal{L_\text{CE}}(\discrmodel(\query_\img[:,N+1,:]), \attr)]
\end{equation}

Furthermore, prior works propose aligning the features across the protected attribute by aligning $p(\hat{\lbl}=y|\attr=0)$ and $p(\hat{\lbl}=y|\attr=1)$. However, this does not take into account the ground truth task label $\lbl$, allowing two datapoints with different task labels to be aligned, which will lead to incorrect classification. Therefore, we propose using \emph{class-specific} alignment where we align the features based on both the protected attribute value \emph{and} the task label value (\emph{i.e.} we only align $p(\hat{\lbl}=y|\attr=0,\lbl=1)$ with $p(\hat{\lbl}=y|\attr=1,\lbl=1)$, and only align $p(\hat{\lbl}=y|\attr=0,\lbl=0)$ with $p(\hat{\lbl}=y|\attr=1,\lbl=0)$). 

\vspace{-.3cm}
\subsubsection{Direct Alignment of Class-Specific Query Activations}
\vspace{-0.1cm}
\label{subsec:queryloss}
Our hypothesis is that by reducing the differences between the average query activations, we can encourage invariance to the protected attribute and improve fairness. Following this we add a Maximum Mean Discrepancy (MMD) loss~\cite{gretton2008kernel} to our objective function that minimizes the L$_2$ distance between the average query activations for instances from each attribute. We exclude the class token of the query activation, $\query_\img[:,N+1,:]$, as it does not encode spatial information. We first compute the average query activation in each minibatch, \batch, for each task label (y) and protected attribute (a) combination, where $\query_{\img_i}$ indicates the query activation for image $\img_i$: 
\vspace{-10pt}
\begin{equation} \label{eq:qavg}
\hat{\query}_{\lbl, \attr} = \frac{1}{|\batch|}\sum_{i\in \batch,\; \text{s.t.}\; \lbl_i=\lbl,\attr_i=\attr}\query_{\img_i}[:,1:N,:]
\end{equation}

Next, we compute the average of $L_2$ differences across the protected attribute for each element in $\hat{\query}_{\lbl,\attr}$ independently, and then average these differences across the $\channel$ channels and $\numHead$ attention heads of the matrix. Due to the reasoning provided in Sec.\ref{subsec:advloss}, we utilize class-specific alignment of the query loss to ensure that the appropriate features are aligned. For binary attributes this yields the following alignment loss: 

\begin{equation} \label{eq:qlcs}
\mathcal{L}_q(\img; \encmodel) = \frac{1}{2\cdot \numHead \channel} \sum_{\attnI \in \numHead}\sum_{\channelI \in \channel} \sum_{\lbl \in \{0,1\}} ||\hat{\query}_{\lbl,0}[\attnI,:,\channelI] - \hat{\query}_{\lbl,1}[\attnI,:,\channelI]||_2
\end{equation}

\noindent Altogether, our model results from the following joint minimization: 

\begin{equation} \label{eq:obj}
\mathcal{L}(\img, \lbl; \encmodel,\clsmodel,\discrmodel) = \mathcal{L}_\text{CE}(\clsmodel(\encmodel(\img)),\lbl)
+ \alpha \mathcal{L}_\text{adv}(\img; \discrmodel, \encmodel)
+ \beta \mathcal{L}_q(\img; \encmodel)
\end{equation}

\noindent where hyper-parameters $\alpha$ and $\beta$ are tuned and optimized for fairness.
\vspace{-5pt}
\section{Experiments}
\vspace{-5pt}

\subsection{Setup \& Implementation Details}
\label{subsec:setup}
\vspace{-5pt}

\begin{figure}[t]
    \centering
    \includegraphics[width=0.65\textwidth]{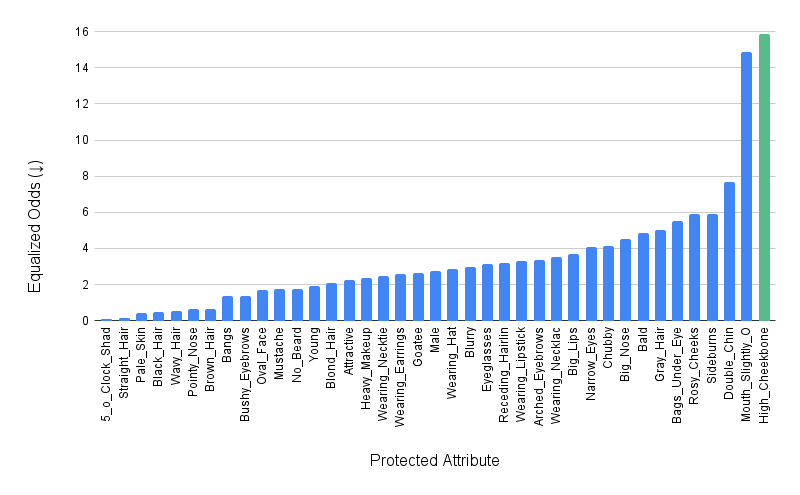}
    \vspace{-15pt}
    \caption{We choose our evaluation settings by measuring the fairness of a Transformer model trained to predict a given task attribute across different protected attributes. 
    This figure above shows the fairness (measured by \eo) across each protected attribute for a Transformer trained to predict Smiling. 
    We select High Cheekbones (in green) as it is the attribute with the greatest bias. A similar process is followed for the (Wavy Hair, Male) and (Wavy Hair, Wearing Lipstick) settings.}
    \vspace{-5pt}
    \label{fig:tpameasures}
\end{figure}

We evaluate \method on CelebA~\cite{liu2015deep}, a dataset containing 200k celebrity face images with annotations for 40 binary attributes. We present results on three settings, each with a corresponding binary task ($\lbl$) that the model is trained to predict, and a binary protected attribute ($\attr$) over which we wish the model to be unbiased. The three settings described as a tuple ($\lbl,\attr$) are as follows: i) (Smiling, High Cheekbones), ii) (Wavy Hair, Male), and iii) (Wavy Hair, Wearing Lipstick). 
We take careful steps to choose these settings: First we train a baseline Transformer model on different task attributes (including Smiling and Wavy Hair). Then we measure the widely-used \eo fairness metric to estimate how much bias is correlated with each protected attribute (results for Smiling reported in Figure \ref{fig:tpameasures}). Finally, we choose the protected attribute with the highest \eo and thus the most bias (more details in the supplement).
% , which also has a suitable ground truth data distribution
% The resulting 3 settings have a large range of spatial correlations and ground truth distributions because the (Smiling, High Cheekbones) contains attributes that occur in a spatially similar region, (Wavy Hair, Wearing Lipstick) occurs in spatially different regions, and (Wavy Hair, Male) spans several spatial regions.} 

\vspace*{-0.5cm}
\noindent \textbf{Implementation Details.} We first train a transformer from scratch on each of these 3 settings. The transformer consists of 196 patches (each representing a 16x16 area of the image), 1 class token patch, 12 transformer encoder layers, and 8 attention heads. We flatten and project each patch into a 64-dimensional vector and add positional embeddings~\cite{wang2020fairness}. The embedded patches are fed into the transformer encoder. We use parameter sharing across key and value weights~\cite{wang2020linformer}. After the transformer encoder processes the patch embeddings, the class token patch is fed into 2 fully-connected layers and a sigmoid layer to produce a single normalized output score (since we deal with binary classification). We use the Adam optimizer~\cite{kingma2017adam} with a learning rate of  1$\times 10^{-4}$, train for 50 epochs, and following Wang~\emph{et al.}\cite{wang2020fairness}, select the model with the best weighted average precision (AP) on the validation set.

We implement \method by applying a class-specific adversarial and class-specific query loss on top of our pretrained transformer model to perform targeted debiasing. 
We align the query class token activations by applying one adversary head per setting of the task label. 
The adversary takes in the 512-dimensional vector query class token of the final (12$^\text{th}$) layer of the transformer encoder. The class token is processed through 2 fully-connected layers and a sigmoid layer to produce a single normalized output score of the binary protected attribute $\attr$. 
%We align the query activations by adding an L2 loss on the activations of the final query matrix. We take the L2 difference of each element in our query matrix, and then average these differences across the 64 channels and 8 attention heads. We exclude the class token from the L2 loss. 
We perform a sweep across multiple loss weights for the adversarial ($\alpha$) and query ($\beta$) losses, and select the one yielding the best fairness metrics.
%\jh{does the value change per setting?}

\noindent \textbf{Metrics.} As performance metrics, we report \sa and \ba \cite{park2020readme}, $\dfrac{1}{4}[TPR_{\attr=0} + TNR_{\attr=0} + TPR_{\attr=1} + TNR_{\attr=1}]$, which is needed when task data is imbalanced.
\noindent As fairness metrics, we report \eo \cite{hardt2016equality} and  introduce \bad. \eo is used to understand the discrepancy in true positive rates and false positive rates across the protected attribute. \bad looks at the difference in \ba between the binary values of the protected group. By computing the difference in \ba, as opposed to difference in \sa proposed by Dieterich~\emph{et al.}\cite{dieterich2016compas}, we are able to account for the class imbalance in the original dataset. 
%This measure is important because it provides a more holistic understanding of the Equalized Odds metric. 
A detailed use case is described in the supplement.
\begin{equation} \label{eu_eqn}
 \text{Balanced\ Acc.\ Difference} (\Delta \text{BA}) = \dfrac{1}{2}[TPR_{\attr=0} + TNR_{\attr=0}] - \dfrac{1}{2}[TPR_{\attr=1} + TNR_{\attr=1}]
\end{equation}
%Imagine a situation wherein after debiasing a model, the tpr difference across a protected attribute increases slightly while fpr decreases substantially. Since Eq Odds is an avg of these two numbers, the resulting measure will be low, indicating that the model is performing relatively strong in terms of fairness compared to the original model. However, due to the high tpr difference, predicting the positive outcome for the protected attribute is actually more unfair than before debiasing, which means that a large bias towards predicting the positive outcome for the protected attribute. In this situation, the Bal Acc Diff measure will be high because this measure looks at the differences in accuracy between a subgroup. Therefore, a user will understand that the drop in Equalized Odds does not tell the full story and that their model still encodes a bias.

\vspace{-20pt}
\subsection{Baselines}
\label{subsec:Baselines}
\vspace{-5pt}
We evaluate the following debiasing algorithms with the pretrained transformer as baselines.

\noindent \textbf{i) Maximum Mean Discrepancy (MMD) ~\cite{long2015learning}: }
MMD computes a mean of penultimate layer feature activations for each setting of the protected attribute, and then minimizes their L$_2$ distance. We select a loss weight that yields the highest validation results.

\noindent \textbf{ii) Domain Adversarial Neural Network (DANN)~\cite{ganin2015unsupervised}:}
%Adversarial learning has been successfully applied as a debiasing strategy in prior work~\cite{hardt2016equality}.
Following successful adversarial debiasing~\cite{hardt2016equality}, we build on domain adversarial alignment~\cite{ganin2015unsupervised} using an attribute adversary learned on top of the penultimate layer activations.
The adversarial head consists of 2 linear layers that take in a 512-dimensional vector class token, followed by a sigmoid, and select the adversarial loss weight that yields the highest validation results.

\noindent \textbf{iii) LAFTR~\cite{madras2018learning}:} We train a model with a modified adversarial objective which attempts to satisfy the Equalized Odds fairness measure. This objective is implemented by minimizing the average absolute difference on each task attribute-protected attribute ($\lbl,\attr$) combination. This allows the model to directly optimize for the equalized odds measure. 

\noindent \textbf{iv) Domain Independent Training~\cite{wang2020fairness}: }
We learn a shared feature representation with an ensemble of classifiers (two), trained with the same baseline hyperparameters. Given the predictions, we perform inference by averaging the class decision boundaries. 
%Domain independent training~\cite{wang2020fairness} learns two separate classification heads corresponding to each value of the protected attribute. Note that we train a separate transformer from scratch for this method with two classification heads, using identical hyperparameters.

% \subsection{Implementation Details}

% \textbf{DANN Lin Atop Q:}
% We compare \method to all 3 previous benchmarks. 

\vspace{-15pt}
\subsection{Results}
\vspace{-5pt}

\begin{table}
  \setlength{\tabcolsep}{4pt}
  \resizebox{\textwidth}{!}{
  \begin{tabular}{lcccccccccccc }
      \toprule
       \multirow{2}{*}{\textbf{Method}} &\multicolumn{4}{c}{\textbf{Y:} Smiling \textbf{A:} High Cheekbones} & \multicolumn{4}{c}{\textbf{Y:} Wavy Hair \textbf{A:} Male} & \multicolumn{4}{c}{\textbf{Y:} Wavy Hair \textbf{A:} Wearing Lipstick} \\
       \cmidrule(l{4pt}r{4pt}){2-5}
       \cmidrule(l{4pt}r{4pt}){6-9}
       \cmidrule(l{4pt}r{4pt}){10-13}
      & \small{EO $\downarrow$} & \small{$\Delta$ BA ($\%) \downarrow$}  & \small{BA (\%)$\uparrow$} & \small{Acc (\%)$\uparrow$} & \small{EO $\downarrow$} & \small{$\Delta$ BA ($\%) \downarrow$}  & \small{BA (\%)$\uparrow$} & \small{Acc (\%)$\uparrow$} & \small{EO $\downarrow$} & \small{$\Delta$ BA ($\%) \downarrow$}  & \small{BA (\%)$\uparrow$} & \small{Acc (\%)$\uparrow$}\\ 
      \midrule            
      Transformer & 15.89 & 9.18 & 87.00 & 92.21 & 19.76 & 12.10 & 72.79 & 78.18 & 19.72 & 9.86 & 74.33 & 78.81 \\      
      \texttt{MMD}~\cite{long2015learning} & 14.83 & 3.26 & 87.65 & 92.61 & 20.86 & 12.26 & 72.90 & 78.51 & 17.88 & 7.82 & 76.00 & \textbf{79.98} \\      
      
      \texttt{DANN}~\cite{ganin2015unsupervised} & 15.13 & 3.58 & 87.67 & \underline{92.72} & 19.74 & 9.36 & 74.70 & 79.65 & 18.63 & 7.67 & 75.21 & 79.31 \\      
      \texttt{LAFTR}~\cite{madras2018learning} & 16.13 & 4.95 & 87.42 &  \textbf{92.78} & 19.03 & 9.24 & 74.75 & 79.55 & 18.17 & \textbf{6.50} & 76.11 & 79.55 \\      
      % \rowcolor{Gray} 
      \texttt{Domain Ind.}~\cite{wang2020fairness} & \textbf{13.12} & 8.59 & 86.61 & 90.89 & \textbf{14.99} & 11.01 & 69.71 & 74.09 & \textbf{13.29} & 9.03 & 70.61 & 73.80 \\      
       \method (Ours) & \underline{14.77} & \textbf{2.42} & \textbf{87.73} & 92.68 & \underline{18.58} & \textbf{8.63} & \textbf{75.10} & \textbf{79.73} & \underline{17.31} & \underline{7.08} & \textbf{76.11} & \underline{79.91} \\
      \bottomrule
      \end{tabular}}
      \vspace{-5pt}
      \caption{{\small Debiasing results. Y=Task Attribute. A=Protected Attribute. EO=Equalized Odds. $\Delta$BA=Balanced Accuracy Difference. BA=Balanced Accuracy. Best and second best performing method bolded and underlined respectively.}}
      % \vspace{-5pt}
      \label{table:TransfSHCB}
\end{table}

We report results in Table~\ref{table:TransfSHCB}. We find that \method consistently improves both \eo and \bad while maintaining \ba and \sa compared to the original transformer model (best or second best in 11/12 settings). 
Within a test setting there may be other debiasing methods which produce stronger debiasing results according to a single fairness metric, such as LAFTR on the (Wavy Hair,Wearing Lipstick) setting, but we note that across settings and considering all metrics our method consistently performs comparably or better than prior work. We further note that while Domain Independent training provides the strongest \eo performance, this approach does not produce strong \bad (fairness) or \ba and \sa (performance), likely due to imbalanced task data for the underrepresented protected attribute. For example, in the (Wavy Hair, Male) setting, though we see a 3.59\% drop in \eo compared to \method, this is accompanied by worse performance on all other metrics including: a 2.38\% increase in \bad, a 5.39\% drop in \ba, and a 5.64\% drop in \sa. 
\vspace{-.5cm}
\subsection{Ablating \method}
\vspace{-0.1cm}
\begin{table}[t]
    \resizebox{\textwidth}{!}{
    \begin{tabular}{ccccccccccccccc}
        \toprule
         & \multicolumn{6}{c}{\centering \bf Adversarial Loss}& \multicolumn{2}{c}{\centering \bf Query Loss} & \multicolumn{2}{c}{\centering \bf Fairness Metrics $\downarrow$} & \multicolumn{2}{c}{\centering \bf Performance Metrics $\uparrow$} & \\
        \cmidrule(l{4pt}r{4pt}){2-7}
        % \cmidrule(l{4pt}r{4pt}){4-5}
        \cmidrule(l{4pt}r{4pt}){8-9}
        \texttt{\#} &  
        \multicolumn{2}{c}{\centering activations} &
        \multicolumn{2}{c}{\centering on top of} & \multicolumn{2}{c}{\centering alignment} &   \multicolumn{2}{c}{\centering alignment} & \\
        \cmidrule(l{4pt}r{4pt}){2-3}
        \cmidrule(l{4pt}r{4pt}){4-5}
        \cmidrule(l{4pt}r{4pt}){6-7}
        \cmidrule(l{4pt}r{4pt}){8-9}
        \cmidrule(l{4pt}r{4pt}){10-11}
        \cmidrule(l{4pt}r{4pt}){12-13}
        &  {\centering penultimate} & {\centering query } & {\centering full }&  {\centering class token } & {\centering full } & {\centering class spe } & {\centering full } & {\centering class spe } & {\centering Eq Odds }& {\centering $\Delta$ Bal Acc } & {\centering Bal Acc} & {\centering Std Acc}\\
        \toprule                 
        1* \cite{dosovitskiy2020image}& \multicolumn{8}{c}{  \centering (Original Transformer Model)} & 15.89 & 9.18 & 87.00 & 92.21 \\
        % \midrule
        2* \cite{ganin2015unsupervised} & \ding{51} &  & \ding{51} &  & \ding{51} &  &  &  & 15.13 & 3.58 & 87.67 & 92.72 \\
        \midrule         
        3 & \ding{51} &  & \ding{51} &  &  & \ding{51} &  &  & 14.64 & 3.46 & 88.01 & 92.90  \\
        % \midrule                    
        4 &  & \ding{51}  & \ding{51} &  & \ding{51} &  &  &  & 15.35 & 3.54 & 87.71 & 92.84 \\           
        5 &  & \ding{51}  & \ding{51} &  &  & \ding{51} &  &  & 14.93 & 3.89 & 87.59 & 92.57 \\           
        6 &  & \ding{51} &  & \ding{51}  & \ding{51} &  &  & & 16.17 & 2.24 & 87.24 & 92.67 \\           
        7 &  & \ding{51} &  & \ding{51} &  & \ding{51} &  &  & 14.34 & 2.98 & 87.90 & 92.69  \\                    
        8 &  & \ding{51} &  & \ding{51} &  & \ding{51} & \ding{51} &  & 15.10 & 2.40 & 87.48 & 92.55  \\   
        \midrule
        9 (\method) &  & \ding{51} &  & \ding{51} &  & \ding{51} &  & \ding{51} & 14.77 & 2.42 & 87.73 & 92.68  \\  
        \bottomrule
        \end{tabular}}
        \vspace{-2pt}
        \caption{Ablating \method. This analysis portrays the importance of applying a class specific adversary to align the class token activations across the protected attribute, and performing MMD alignment on the query matrix activations across the protected attribute. * indicates a baseline method we are comparing against.}\vspace{-5pt}
        \label{tab:ablations}
\end{table}

In Table~\ref{tab:ablations} we ablate \method and observe that each component of our method contributes to improved fairness measures. %First, applying class-specific alignment is more effective than class agnostic for penultimate (row 2 vs 3) and query features (row 4 vs 5, 6 vs 7, and 8 vs 9).
\setlist[itemize]{leftmargin=*}
\begin{itemize}
\renewcommand{\labelitemi}{$\Rightarrow$}
\itemsep-0.5em 
\item \textbf{Class specific adversary improves fairness (Row 3):} We first experiment with applying a class-specific adversarial loss to align distributions within a setting of the task attribute. This leads to improvements in both fairness metrics and both accuracy metrics.  
\item \textbf{Positioning adversary head on full query activations is ineffective (Row 4 \& 5)}: Next we move the adversary head on top of the full output of the final query matrix. We experiment with both a standard (Row 4) and class-specific (Row 5) adversarial loss. While this does not do better than DANN, we hypothesize that this is because the input to the adversary is the entire query matrix rather than just the class token, which is what is actually used for task classification.
\item \textbf{Only operating on the query class token greatly improves fairness(Row 6 \& 7)}: To test our prior hypothesis, we allow the adversary to only operate on the class token of the query matrix. We experiment with both a standard (Row 6) and class specific (Row 7) adversarial loss and notice that the latter results in strong \eo measures.
\item \textbf{Adding query loss improves \bad but sacrifices \eo (Row 8):} Next, following our intuition of equalizing the query activations, we add a query loss. We notice an improvement in \bad but no significant improvement in \eo. We again hypothesize that class specificity is required to notice improvements. 
\item \textbf{Adding class specificity to query loss improves \eo while maintaining \bad(Row 9)}: Finally, a class-specific query loss combined with a class-specific adversarial loss on the query class token achieves the strongest and most consistent performance across all metrics, implying that each component of \method is helpful in reducing bias while maintaining strong predictive performance.
\end{itemize}

\vspace{-0.7cm}
\subsection{Analysis}
\begin{figure}[t]
\centering
%% \vspace{-10pt}
\includegraphics[width=0.60\textwidth]{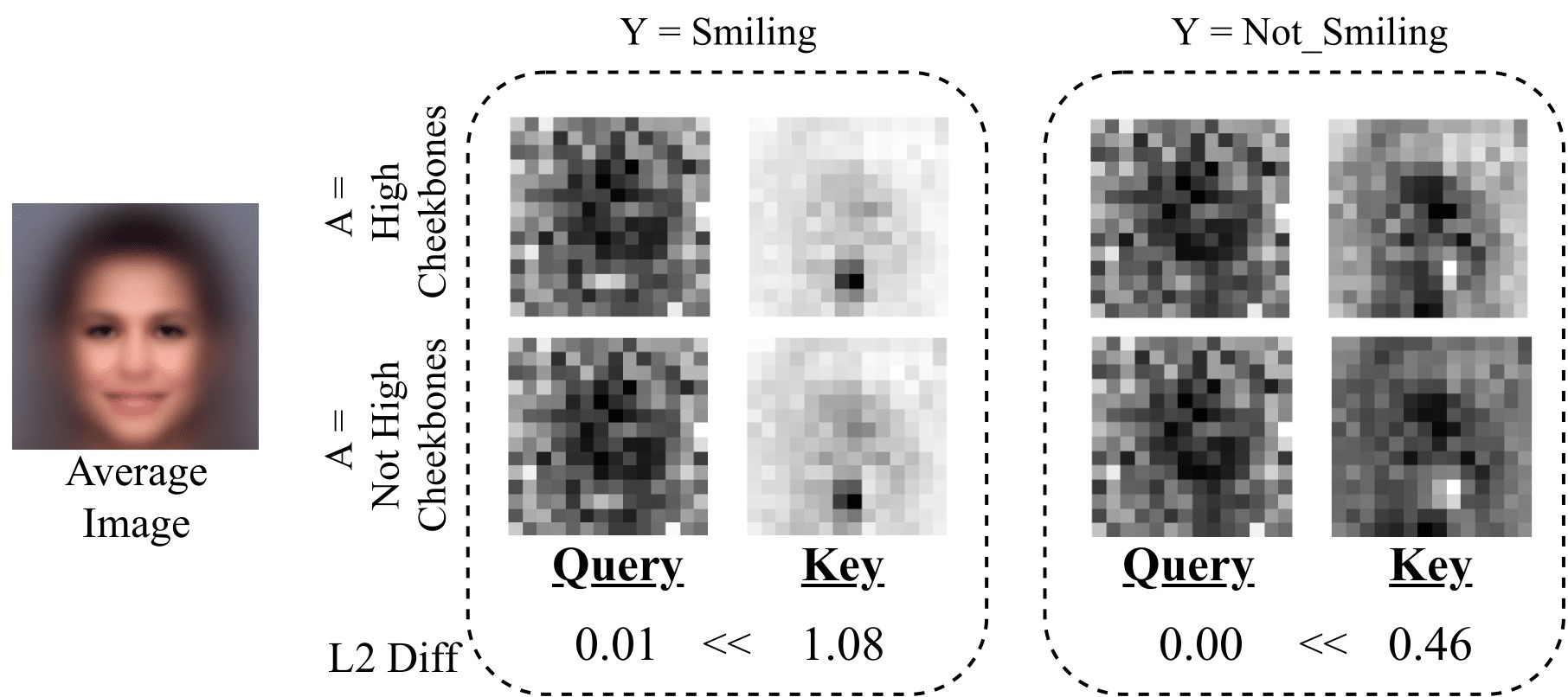}
\caption{We show the average query and key activations for a fixed task label (Y=smiling or not smiling) and a fixed attribute label (A = High Cheekbones or not High Cheekbones) on the CelebA dataset after debiasing. The difference in query activations across attribute labels is reduced compared to the results observed before debiasing (Figure \ref{fig:qmbefore}).}
\label{fig:qmafter}
\end{figure}
\vspace{-7pt}
As seen in Figure~\ref{fig:qmbefore} (\emph{left}), before applying \method, the average query activations across both settings of the protected attribute are very different (large L$_2$ distance of 5.82 and 2.4 for smiling and not smiling respectively), while the average activations from the key matrix are much closer (L$_2$ distance of 3.07 and 0.49). 
These qualitative visualization results and quantitative L$_2$ results are consistent across different attention heads and channels of the query and key activations (additional results provided in the supplement).
%

%there is a bias being learned primarily by the query matrix that does not exist as significantly in the key matrices, and to debias the model, we must train the query matrix to be invariant to the protected attribute. Following this hypothesis, we focus on the query matrix as a basis of the bias mitigation algorithm we present in Sec.~\ref{subsec:tadot}. (Note that in our implementation, the value matrix weights are shared with the keys, therefore we do not focus on visualizing the value matrices as their activations are very similar to the keys~\cite{wang2020linformer}).\\

After performing debiasing using \method, we visualize the new values of the average query and key activations and calculate the same L$_2$ distances. As seen in Figure \ref{fig:qmafter}, the query activations look very similar and have a small L$_2$ difference (0.01 and 0.0) across settings of the protected attribute. These results, combined with our improved fairness results with \method (Table \ref{table:TransfSHCB}), validates our hypothesis that targeted debiasing focused on the query activations is an effective debiasing strategy for visual transformers.

\noindent 
\textbf{Comparing CNN's and Transformers.} We train a CNN on the same 3 settings described previously, and evaluate accuracy and fairness. As shown in Tab. \ref{table:cnnvstransf}, the baseline Transformer consistently performs significantly worse on both fairness metrics (\eo and \bad). However, we refrain from concluding that Transformers are more biased than CNN's, because we see that Transformers also perform slightly worse in terms of accuracy measures (\ba and \sa), which makes this an unfair comparison. As techniques develop to improve the accuracy of Vision Transformers to be commensurate with CNNs, we hope to address this question more conclusively.
\vspace{-7pt}
\begin{table}[h]
  \setlength{\tabcolsep}{4pt}
  \resizebox{\textwidth}{!}{
  \vspace{-10pt}
  \begin{tabular}{lcccccccccccc }
      \toprule
       \multirow{4}{*}{\textbf{Method}} 
       & \multicolumn{4}{c}{\textbf{Y:} Smiling \textbf{A:} High Cheekbones} & \multicolumn{4}{c}{\textbf{Y:} Wavy Hair \textbf{A:} Male}  & \multicolumn{4}{c}{\textbf{Y:} Wavy Hair \textbf{A:} Wearing Lipstick} \\
       \cmidrule(l{4pt}r{4pt}){2-5}
       \cmidrule(l{4pt}r{4pt}){6-9}
       \cmidrule(l{4pt}r{4pt}){10-13}
      & \small{EO $\downarrow$} & \small{$\Delta$ BA ($\%) \downarrow$}  & \small{BA (\%)$\uparrow$} & \small{Acc (\%)$\uparrow$}  & \small{EO $\downarrow$} & \small{$\Delta$ BA ($\%) \downarrow$}  & \small{BA (\%)$\uparrow$} & \small{Acc (\%)$\uparrow$} & \small{EO $\downarrow$} & \small{$\Delta$ BA ($\%) \downarrow$}  & \small{BA (\%)$\uparrow$} & \small{Acc (\%)$\uparrow$}  \\ 
      \midrule            
      CNN & 14.66 & 2.69 & 88.15 & 93.06 & 16.71 & 8.08 & 77.99 & 82.20 & 15.64 & 6.41 & 78.75 & 82.20 \\
      Transformer & 15.89 & 9.18 & 87.00 & 92.21 & 19.76 & 12.10 & 72.79 & 78.18 & 19.72 & 9.86 & 74.33 & 78.81\\
      
      \bottomrule
      \end{tabular}}
      \vspace{-10pt}
      \caption{{\small Fairness and Accuracy metrics compared for a CNN and Transformer trained on the same task attribute-protected attribute combination. Y=Task. A=Protected Attribute. EO=Equalized Odds. $\Delta$BA=Balanced Accuracy Difference. BA=Balanced Accuracy. }}
      % \vspace{-5pt}
      \label{table:cnnvstransf}
\end{table}
% \vspace{-cm}
\vspace*{-35pt}
\section{Conclusion}
\vspace{-10pt}

In this work, we perform the first benchmarking of several existing debiasing algorithms on visual transformers. We visualize the feature space learned by the transformer self-attention modules and find a significant portion of the bias is encoded in the query matrices. Using this information, we introduce \method, a targeted debiasing algorithm for transformers that aims to remove bias encoded within the query matrix.

Our method suffers from some limitations. First, we note that it is important for users to consider the application of their model before using our findings. In this work, we measure fairness using \eo and \bad, but if one's intended use case does not align with these definitions of fairness, the results presented may not be applicable. Furthermore, all previous methods benchmarked in this paper, along with \method, require annotations for the protected attribute, which may not always be feasible.

\noindent\textbf{Acknowledgments:} This work was funded in part by Cisco Inc.
 
\bibliography{egbib}
\newpage
\section{Supplementary Work}
\subsection{Evaluation Setting Details}
\label{subsec:settings}

In Sec 4.1 we mention that we choose three settings to benchmark prior work and evaluate \method. These three settings, described as a tuple ($\lbl,\attr$), are as follows: i) (Smiling, High Cheekbones), ii) (Wavy Hair, Male), and iii) (Wavy Hair, Wearing Lipstick). Here we provide details on the choice of these three settings.
\subsubsection{Smiling with High Cheekbones}
\begin{wrapfigure}{R}{0.4\textwidth}
  \centering
  \vspace{-20pt}
  \includegraphics[width=\linewidth]{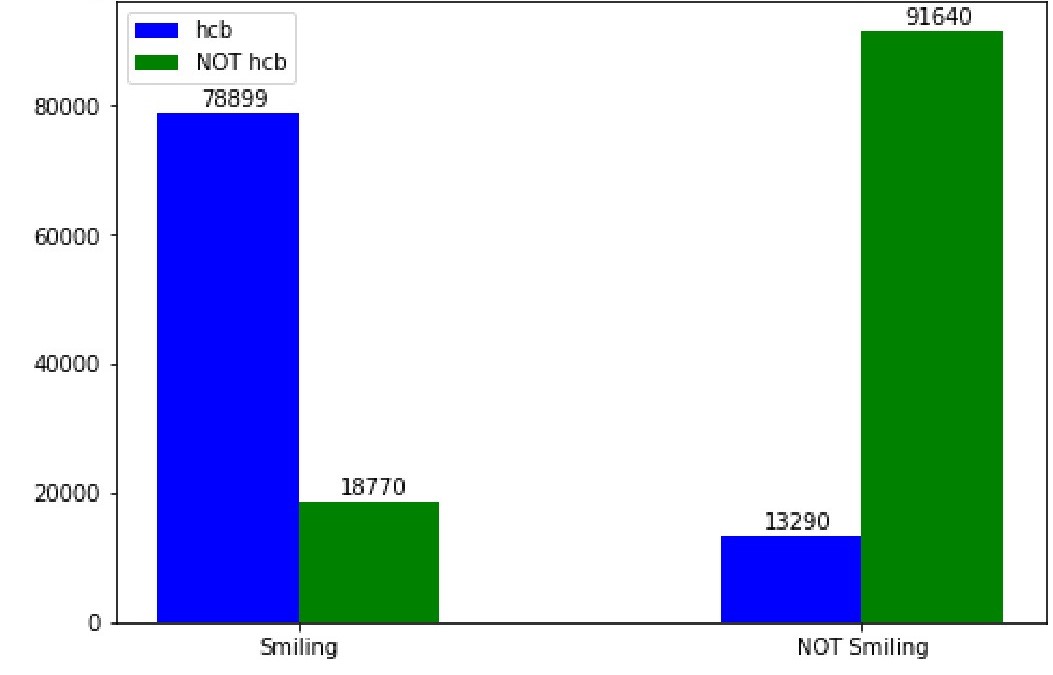}
      \caption{Full dataset distribution of task, protected attribute tuple (Smiling, High Cheekbones).}
      \vspace{-12pt}
      \label{fig:shcbd}
\end{wrapfigure}
The first setting we evaluate on is (Smiling, High Cheekbones) (See Fig \ref{fig:shcbd}). We notice that there is a significant skew in the distribution of the CelebA dataset, where most ``Smiling'' faces are correlated with ``High Cheekbones'' and most ``Not Smiling'' faces are correlated with ``Not High Cheekbones''. However, we understand that not all such correlations will necessarily translate to a \emph{model} bias. Therefore, we analyze the true positive rate and false positive rate of a transformer trained on this dataset to predict the ``Smiling'' attribute, for each setting of the protected attribute ``High Cheekbones''. We obtain the following results: $\text{TPR}_{\attr=1} = 92.83\%$, $\text{TPR}_{\attr=0} = 67.76\%$, $\text{FPR}_{\attr=1} = 9.65\%$, and $\text{FPR}_{\attr=0} = 2.94\%$. Clearly, the model performs significantly worse on correctly predicting ``Smiling'' when the individual does \emph{not} have ``High Cheekbones'' ($\attr=0)$. Furthermore, the model has a $6.71\%$ larger FPR for the ``High Cheekbones'' group ($\attr=1)$. This confirms that the model indeed has a strong bias of (spuriously) correlating the presence of ``High Cheekbones'' with whether they are ``Smiling''. Due to this clear bias, we choose (Smiling, High Cheekbones) as our first setting for evaluation.
\subsubsection{Wavy Hair with Male}
\label{subsec:whm}
\begin{wrapfigure}{R}{0.4\textwidth}
  \centering
  \vspace{-20pt}
  \includegraphics[width=\linewidth]{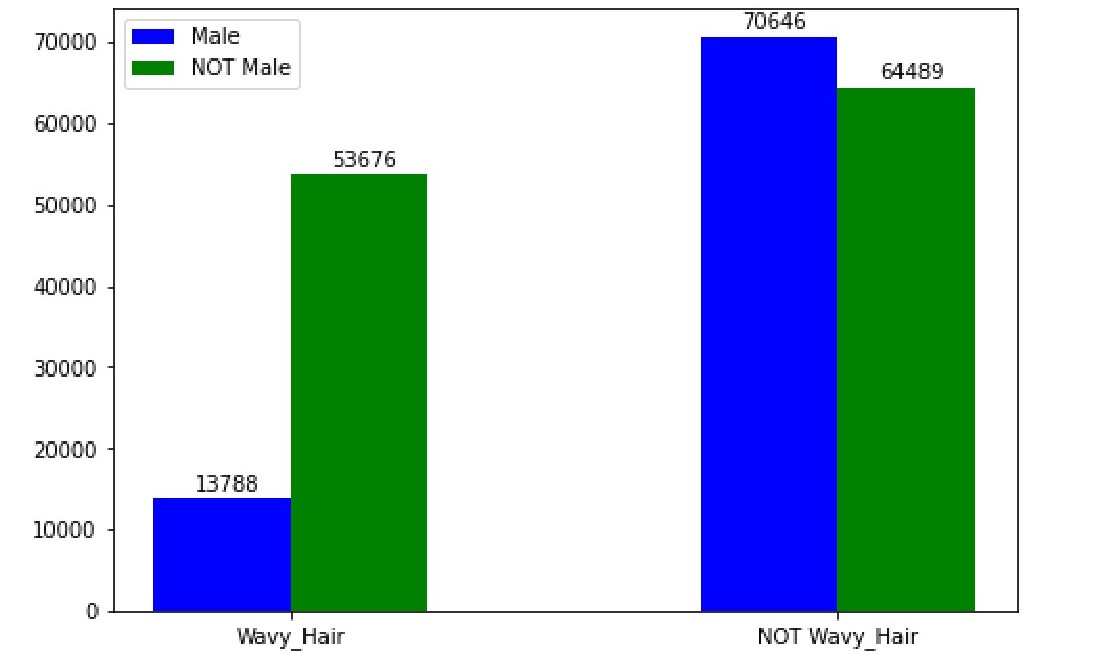}
      \caption{Full dataset distribution of task, protected attribute tuple (Wavy Hair, Male).}
      \vspace{-12pt}
      \label{fig:wmd}
\end{wrapfigure}
Next, we evaluate on (Wavy Hair, Male) (See Fig \ref{fig:wmd}). First, we notice that the dataset contains less ``Male'' individuals (28.5\%) than ``Not Male'' individuals (71.5\%). Due to the fact that ``Male'' is underrepresented in the data, we suspect that the model will not be able to learn as strong of a representation of the images with the ``Male'' attribute as it will for the ``Not Male'' attribute, which could lead to some type of bias. Next, we notice that most ``Male'' individuals are labeled as ``Not Wavy Hair'', which could cause a spurious correlation between these two attributes. When analyzing a Transformer trained on the ``Wavy Hair'' prediction task, we notice exactly that: a high difference in TPR where $\text{TPR}_{\attr=1} = 36.06\%$ and $\text{TPR}_{\attr=0} = 67.92\%$. This indicates that a (spurious) dataset correlation of ``Male'' with ``Not Wavy Hair'' is being learned by the model, leading to a large bias wherein the model is overpredicting ``Male'' to have ``Not Wavy Hair''. 
\subsubsection{Wavy Hair with Wearing Lipstick}
\begin{wrapfigure}{R}{0.4\textwidth}
  \centering
  \vspace{-20pt}
  \includegraphics[width=\linewidth]{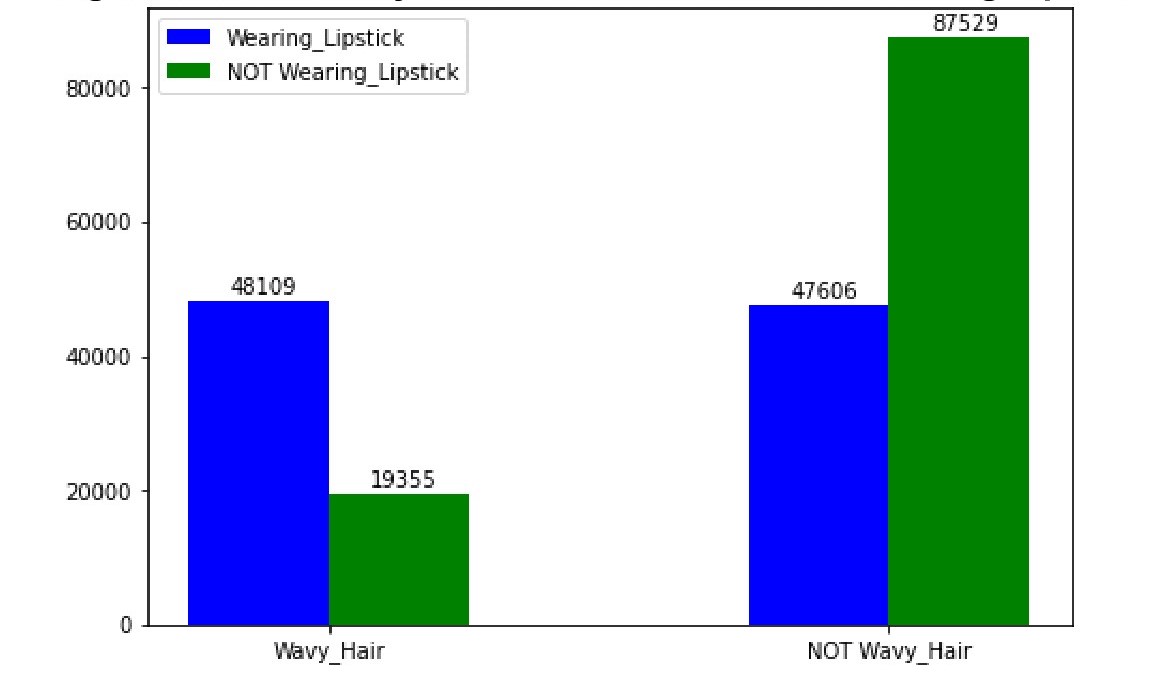}
      \caption{Full dataset distribution of task, protected attribute tuple (Wavy Hair, Wearing Lipstick).}
      \vspace{-12pt}
      \label{fig:wwld}
\end{wrapfigure}
Finally, we evaluate on (Wavy Hair, Wearing Lipstick) (See Fig \ref{fig:wwld}). There is a natural correlation in the dataset where most ``Male'' individuals are ``Not Wearing Lipstick'', and vice versa. Since ``Wearing Lipstick'' and ``Male'' are correlated, and ``Male'' and ``Wavy Hair'' are correlated (as shown in \ref{subsec:whm}), we want to see if the bias that is present when the protected attribute is ``Male'' will persist when we set the protected attribute as ``Wearing Lipstick''. Looking at the dataset distributions themselves, we notice that the number of people ``Wearing Lipstick'' and ``Not Wearing Lipstick'' is closer to 50/50 than in the previous setting (protected attribute=``Male''). Next, we notice that there is a correlation in the dataset of people with ``Wavy Hair'' and people ``Wearing Lipstick''. After training a transformer, we notice that the \eo and \bad are almost as high as the (Wavy Hair, Male) setting. Therefore, we conclude that this setting will be a good test-bed, as the distribution is not as clearly skewed as (Wavy Hair, Male), but there is poor performance on both fairness metrics \eo and \bad.

\subsection{\bad Use Cases}
\label{subsec:baduc}
In Sec. 4.1 of the main paper, we introduce \bad, a fairness metric that shares some of the motivation behind accuracy equity~\cite{dieterich2016compas}, but with an important implementation difference to account for real-world data distributions. While accuracy equity suggests taking the difference in \sa across the protected attribute, we take the difference in \ba (our performance metric) across the protected attribute. By doing so, we can account for class imbalance in the dataset, as we saw in Sec. \ref{subsec:settings}. Furthermore \bad is important because it provides a more holistic understanding of the \eo metric, as we now elaborate. 

Consider a situation wherein after debiasing a model, the true positive rate (TPR) difference across a protected attribute \emph{increases} slightly while false positive rate (FPR) difference decreases substantially. Since \eo is an average of TPR difference and FPR difference ($\text{\eo} = \dfrac{1}{2}[TPR_{\attr=1} - TPR_{\attr=0}] + \dfrac{1}{2}[FPR_{\attr=1} - FPR_{\attr=0}]$), the resulting \eo measure will reduce, indicating that the model is fairer than the original model. However due to the \emph{increased} TPR difference, predicting the positive outcome for the protected attribute is actually \emph{more} unfair than before debiasing! This means that for the positive outcome, a large bias across the protected attribute still exists, which the \eo metric does not adequately capture. 

However, in this situation, the \bad will be high as it looks at the \emph{differences} in \ba within each subgroup. More specifically, \bad can be rewritten as:
\begin{equation} \label{pt0}
 \text{Balanced\ Acc.\ Difference} (\Delta \text{BA}) = \dfrac{1}{2}[TPR_{\attr=0} + TNR_{\attr=0}] - \dfrac{1}{2}[TPR_{\attr=1} + TNR_{\attr=1}]
\end{equation}
\begin{equation} \label{pt1}
 = \dfrac{1}{2}[TPR_{\attr=1} + TNR_{\attr=1}] - \dfrac{1}{2}[TPR_{\attr=0} + TNR_{\attr=0}]
\end{equation}
\begin{equation} \label{pt2}
= \dfrac{1}{2}[TPR_{\attr=1} - TPR_{\attr=0}] + \dfrac{1}{2}[(TNR_{\attr=1} - TNR_{\attr=0}] 
\end{equation}
\begin{equation} \label{pt3}
 = \dfrac{1}{2}[TPR_{\attr=1} - TPR_{\attr=0}] + \dfrac{1}{2}[(1-FPR_{\attr=1}) - (1-FPR_{\attr=0})] 
\end{equation}
\begin{equation} \label{pt4}
 = \dfrac{1}{2}[TPR_{\attr=1} - TPR_{\attr=0}] + \dfrac{1}{2}[(FPR_{\attr=0} - FPR_{\attr=1})] 
\end{equation}
\begin{equation} \label{pt5}
 = \dfrac{1}{2}[TPR_{\attr=1} - TPR_{\attr=0}] - \dfrac{1}{2}[(FPR_{\attr=1} - FPR_{\attr=0})] 
\end{equation}

By Eq.~\ref{pt5}, it is clear that given our situation where TPR difference is slightly higher than before debiasing, and FPR difference is lower than before debiasing, \bad would \emph{increase}, indicating that the model is behaving in a biased manner, even though the \eo measure decreases. Therefore, a user will realize that the drop in \eo does not tell the full story, as \bad will indicate that their model still encodes a bias, especially towards predicting the positive value for one setting of the protected attribute. Hence, we advocate for using \bad as an \emph{additional} fairness metric, along with \eo. 

\subsection{Transformer Feature Visualizations}
\label{subsec:qvk}
\begin{figure}
\centering
\begin{minipage}{.3\textwidth}
  \centering
  \includegraphics[width=1\linewidth]{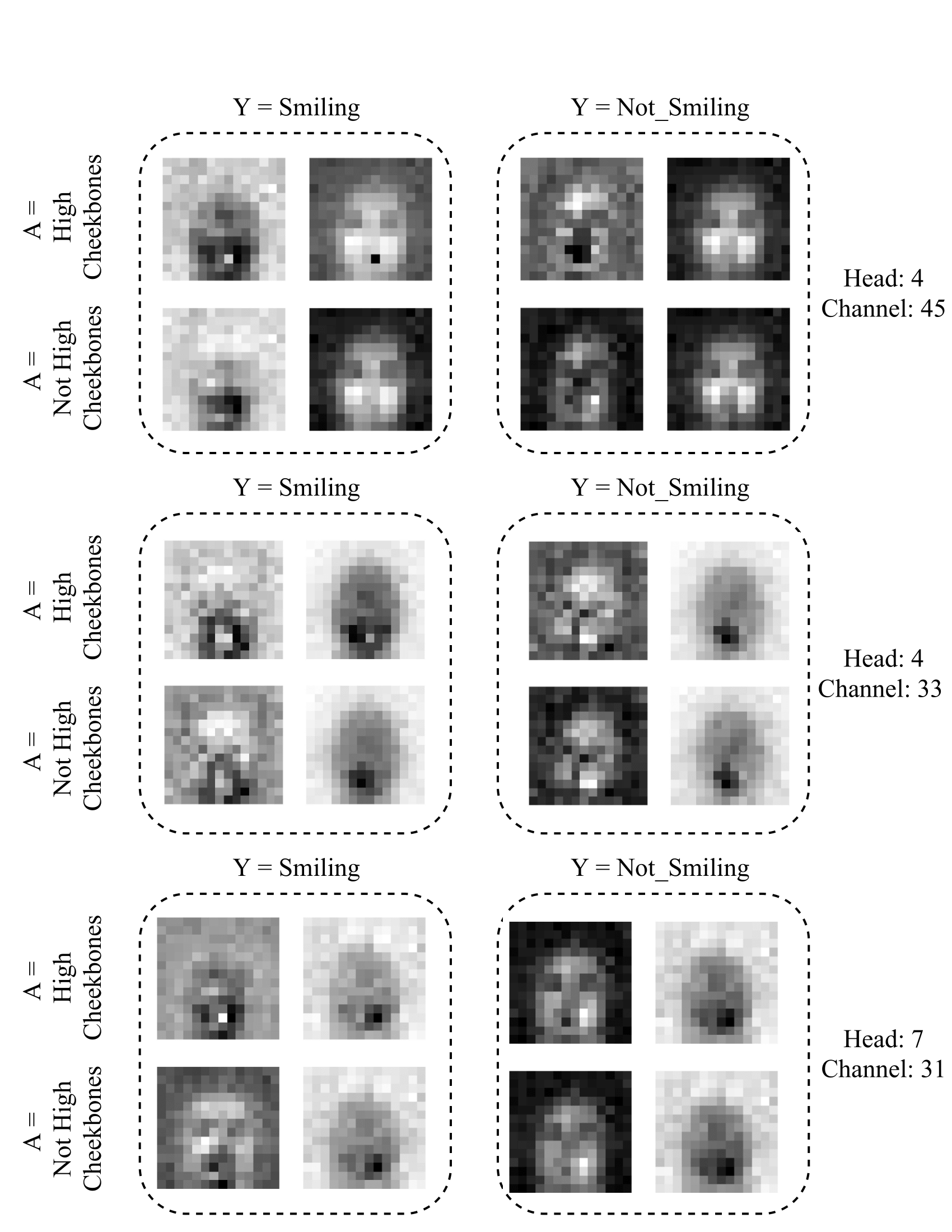}
  \label{fig:sub1}
\end{minipage}%
\begin{minipage}{.3\textwidth}
  \centering
  \includegraphics[width=1\linewidth]{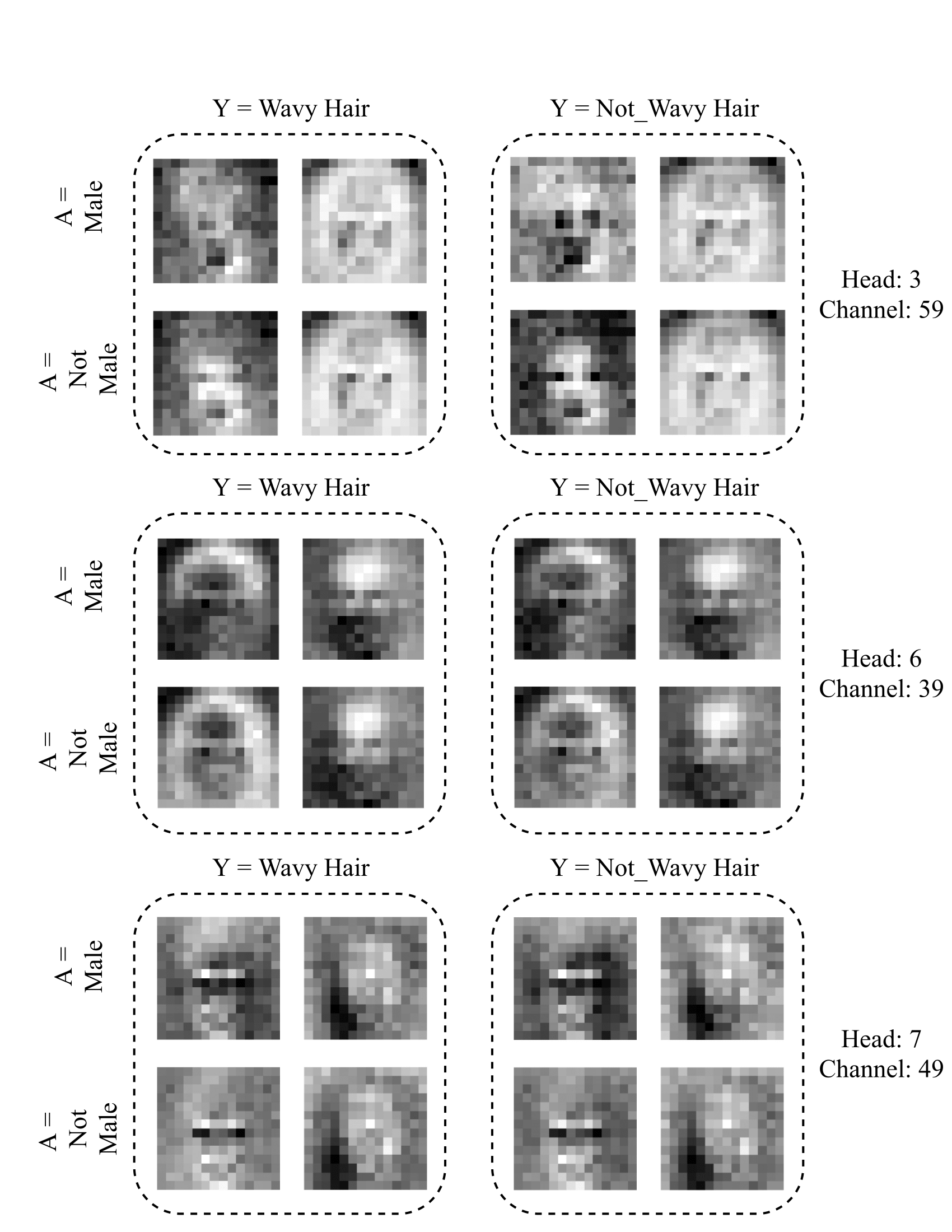}
  \label{fig:sub2}
\end{minipage}
\begin{minipage}{.3\textwidth}
  \centering
  \includegraphics[width=1\linewidth]{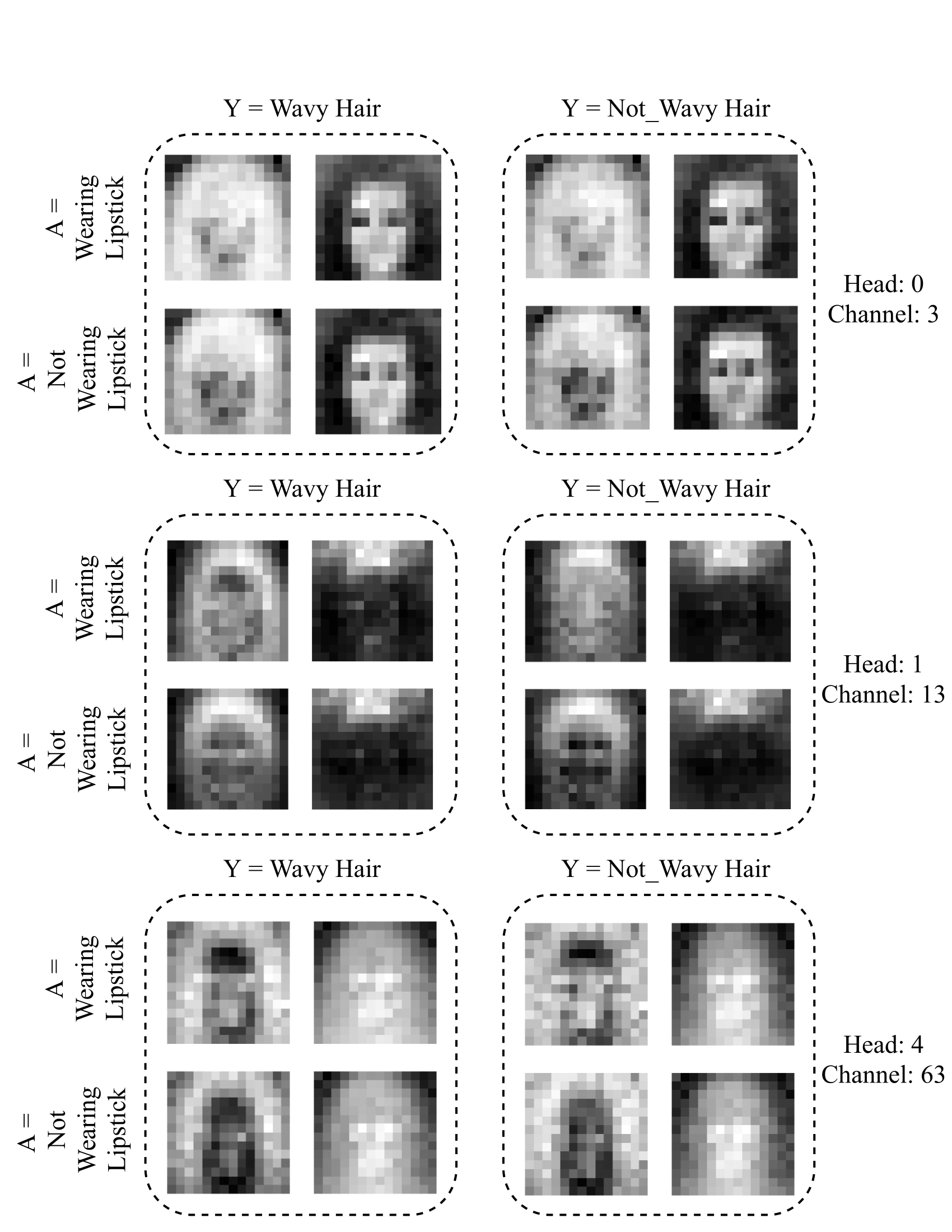}
  \label{fig:sub2}
\end{minipage}
\caption{We show that the variance in the query matrix activations for a particular task label, \lbl, across a protected attribute \attr, generalizes across attention heads, channels, and data settings. The Transformer Query/Key Matrix has 64 channels and 8 heads, and we have chosen 3 random channel/head combinations from all 3 tasks to depict the generalization of differences in activations.}
\label{fig:test}
\end{figure}

Recall that in Figure 1 of the paper, we presented a visualization of the average Query and Key activations for each (\lbl,\attr) tuple combination, for a specific attention head and channel of a transformer trained for the ``Smiling'' prediction task. In Figure ~\ref{fig:test}, we provide visualizations that demonstrate that the differences we notice in the query activations, and the similarity noticed in the key activations, generalizes across different attention heads and channels of the query and key matrices. Further, this also generalizes across different tasks.  
% we provide visualizations that demonstrate this difference across multiple combinations of attention heads, channels, and task settings.

\subsection{Analyzing Class-Specific Alignment for CNNs}
\label{subsec:cnnexp}
In \method, we propose using class-specific alignment wherein we align the distribution of the protected attribute within the task attribute by utilizing an adversary head \emph{per-attribute} during adversarial learning. We have shown the benefits of this method for debiasing visual transformers. In Table \ref{table:CNNSHWM}, we show that such class-specific alignment for adversarial training improves upon previous debiasing algorithms in most settings for CNN's as well.

\begin{table}[h]
  \setlength{\tabcolsep}{4pt}
  \resizebox{\textwidth}{!}{
  \begin{tabular}{lcccccccccccc }
      \toprule
       \multirow{4}{*}{\textbf{Method}} 
       & \multicolumn{4}{c}{\textbf{Y:} Wavy Hair \textbf{A:} Male} & \multicolumn{4}{c}{\textbf{Y:} Smiling \textbf{A:} High Cheekbones} \\
       \cmidrule(l{4pt}r{4pt}){2-5}
       \cmidrule(l{4pt}r{4pt}){6-9}
      & \small{EO $\downarrow$} & \small{$\Delta$ BA ($\%) \downarrow$}  & \small{BA (\%)$\uparrow$} & \small{Acc (\%)$\uparrow$}  & \small{EO $\downarrow$} & \small{$\Delta$ BA ($\%) \downarrow$}  & \small{BA (\%)$\uparrow$} & \small{Acc (\%)$\uparrow$}  \\ 
      \midrule            
      Original CNN & 16.71 & 8.08 & 77.99 & 82.20 & 14.66 & 2.69 & 88.15 & 93.06 \\
      DANN~\cite{ganin2015unsupervised}  & 14.75 & 7.39  & 77.36  & 81.12 & 15.04 & \textbf{1.85} & 87.97 & 93.03 \\
      DANN Class Specific & \textbf{14.57} & \textbf{7.06}  & 77.21  & 80.89 & \textbf{14.51} & 3.37 & 88.40 & 93.24 \\
      
      \bottomrule
      \end{tabular}}
      \vspace{-5pt}
      \caption{{\small CNN Debiasing results. Y=Task. A=Protected Attribute. EO=Equalized Odds. $\Delta$BA=Balanced Accuracy Difference. BA=Balanced Accuracy. }}
      % \vspace{-5pt}
      \label{table:CNNSHWM}
\end{table}
 
\end{document}